\documentclass[sigplan,10pt]{acmart}
\renewcommand\footnotetextcopyrightpermission[1]{}
\usepackage{multirow}
\usepackage{xspace}
\usepackage{makecell}
\usepackage{colortbl}
\usepackage{makecell}
\usepackage{graphicx}
\usepackage{subcaption}
\usepackage{nicefrac}
\usepackage{enumitem}
\usepackage[ruled,lined,linesnumbered]{algorithm2e}

\definecolor{p25color}{HTML}{D4E6F1}
\definecolor{avgcolor}{HTML}{D5F5E3}
\definecolor{p95color}{HTML}{FADBD8}

\AtBeginDocument{%
  }

\newcommand{\para}[1]{\smallskip\noindent{\bf #1}}
\newcommand{\name}{Kairos\xspace}

\begin{document}

\title{\name{}: A Scalable Serving System for Physical AI\vspace{-6pt}}

\author{Yinwei Dai}
\affiliation{%
  \institution{Princeton University}
  \country{}
}

\author{Ganesh Ananthanarayanan}
\affiliation{%
  \institution{Microsoft}
  \country{}
}

\author{Landon Cox}
\affiliation{%
  \institution{Microsoft}
  \country{}
}

\author{Xenofon Foukas}
\affiliation{%
  \institution{Microsoft}
  \country{}
}

\author{Bozidar Radunovic}
\affiliation{%
  \institution{Microsoft}
  \country{}
}

\author{Ravi Netravali}
\affiliation{%
  \institution{Princeton University}
  \country{}
}

\newcommand{\rn}[1]{{\textcolor{blue}{RN: #1}}}
\newcommand{\yinwei}[1]{{\textcolor{orange}{YD: #1}}}
\newcommand{\ga}[1]{{\textcolor{cyan}{GA: #1}}}
\newcommand{\fon}[1]{{\textcolor{green}{XF: #1}}}
\newcommand{\br}[1]{{\textcolor{red}{BR: #1}}}
\newcommand{\gaa}[1]{{\textcolor{cyan}{#1}}}


\settopmatter{printfolios=true, printacmref=false}

\pagestyle{plain}

\begin{abstract}


Physical AI is experiencing rapid growth with frontier foundation models increasing its capabilities across general environments. 
Physical AI tasks are characterized by inference properties that are markedly different from digital AI. They consist of multiple rounds of inference and action execution, generating a chunk of actions in each inference round, and asynchronously interleaving inference and execution.
This makes existing digital AI serving systems unsuited for physical AI; a shortcoming that is critical for enabling their wide adoption, considering their size and the scale of the robot fleets they have to serve.
To fill this gap, we design \name{}, the first multi-robot serving system that makes the generate-execute loop a first-class citizen, with active involvement in the execution phase. Across a wide range of physical AI models and robots, \name{} reduces the average end-to-end task latency by 31.8--66.5\% over state-of-the-art digital AI serving practices, with gains scaling with the robot fleet size.

\end{abstract}

\maketitle

\section{Introduction}
\label{s:intro}

Physical AI -- intelligent systems that perceive and act on the real world --  is undergoing rapid growth. For instance, robotic platforms ranging from dexterous manipulation arms to full-scale humanoids are being deployed across warehouses, factories, hospitals, and households~\cite{amazon_robotics,tesla_optimus,1x,figure,surgical_robot_survey}, with demand pushing toward fleets of hundreds or thousands of robots coordinating on diverse tasks. Underpinning these deployments is a new generation of frontier foundation models, including Vision-Language-Action models (VLAs), Video Action Models (VAMs), and World Action Models (WAMs), that map real-world sensory observations to robot motor commands with unprecedented generality. As physical AI scales in the size of models and number of robots, offloading inference out of the robot, and consolidating robot inference onto shared serving infrastructure becomes essential for cost efficiency~\cite{offload}.

Serving physical AI at scale, however, surfaces a fundamental difference from digital AI (e.g., LLMs for text generation) that existing serving systems are not designed for. In digital AI, inference is {\em decoupled}
from the world: a model generates tokens from a fixed context, and those tokens remain valid regardless of when they are consumed. Physical AI breaks this property. To meet their high control frequencies, physical AI models generate actions in \emph{chunks}, whereby a single inference call produces N actions that the robot executes sequentially while the next inference round runs \emph{asynchronously} in parallel, keeping the robot in motion until it completes its {\em task}. 
However, robots act in a dynamic, changing world, so the observations used for each inference -- e.g., camera images, joint states -- are snapshots that grow stale as time passes and actions are executed, making later actions in each chunk increasingly outdated. This interplay between inference and a changing physical world has two implications for how serving systems must evolve.


First, \emph{serving systems for physical AI must manage a new accuracy-efficiency tradeoff}. The number of generated actions to execute before a new round of inference (i.e., the {\em execution horizon}) directly controls the tradeoff: a shorter horizon replans with fresher observations (and higher accuracy) at proportionally higher serving load, while a longer horizon lowers inference cost but risks executing stale actions (and lowering accuracy). The optimal execution horizon varies across physical AI tasks, robots, and inference rounds within a task as execution demands shift, e.g., free-space motion tolerates a long horizon, while contact-rich grasping does not. Yet today, execution horizons are set statically by developers, conservatively for the worst case. Our results show that this leaves 62\%–87\% of rounds safely capable of executing up to 2.8$\times$ more actions without accuracy loss, hence 2.8$\times$ lower inference load -- substantial efficiency opportunities that no existing serving system captures.


Second, \emph{serving systems must account for execution times when scheduling}. In digital AI, task progress is purely a function of inference times. 
In physical AI, tasks are multi-round and interleave inference with real-world execution phases of the robot that are significant -- often dominant -- 
in end-to-end latency, and inherently heterogeneous across tasks and robots that share infrastructure (0.3s–1.6s for each execution phase at 30Hz control based on the selected execution horizon). This matters because a task's true urgency for when it next needs an inference is determined by its execution duration, not its generation history. Yet existing schedulers, which reason only about accumulated generation time used for inference or queuing delays, are oblivious to this: a task with little generation time may be the longest-running in wall-clock terms if its execution dominates, causing schedulers to systematically prioritize the wrong tasks. Indeed, across our workloads, this misidentification affects 50-93\% of tasks, inflating end-to-end latency across the fleet. 

We present \textbf{\name{}}, the first serving system designed for physical AI. \name{}’s central insight is that efficient physical AI requires the serving system to be a first-class participant in the generate–execute loop, not just a passive inference provider, both within the rounds of each task and across tasks sharing infrastructure. Within each task, \name{} treats the execution horizon as a dynamic, per-round knob managed by the serving system, adapting it in each round to the robot’s current execution context. To do so without additional inference cost, \name{} exploits the inference confidence of each action’s final diffusion update in the chunk. The confidence serves as a fine-grained staleness indicator, identifying the boundary in each round of actions beyond which generated actions are no longer reliable. Across tasks, minimizing end-to-end latency requires reasoning about when each task needs its next inference, which is gated by whichever phase (generation or execution) currently dominates. \name{} addresses this with \emph{wait ratio} -- the fraction of a task's lifetime spent waiting for compute. The wait ratio unifies phase delays into a trackable priority signal to avoid the inversions that plague generation-only schedulers.

We evaluate \name{} across six physical AI models spanning three architecturally distinct families (VLAs, WAMs, and VAMs), five simulation benchmarks (LIBERO, Meta-World, Isaac Lab, RoboTwin 2.0, SIMPLER), and real-robot experiments on a bimanual SO-101 platform.
Compared to FIFO scheduling in existing LLM serving systems (e.g., vLLM~\cite{vllm}) and fairness-based scheduling in multi-round agent serving systems (e.g., Autellix~\cite{autellix}), \name{} reduces average end-to-end task latency by 31.8--66.5\% at peak load in the online serving setting. When serving a dedicated fleet of robots that collectively execute a fixed set of tasks offline, \name{}'s average latency reductions grow from 20.4\% at 10 robots to 42.8\% at 100 concurrent robots. We will open source \name{} post publication.


In summary, \name{} makes the following contributions. {\bf (1)} We quantify the impact of the key parameter, execution horizon, controlling the accuracy-efficiency tradeoff (\S\ref{s:background}) and adapt it dynamically for resource savings (\S\ref{s:design_exec_horizon}). {\bf (2)} We build a scheduling system for a fleet of robots designed for the workload pattern of generation-execution rounds (\S\ref{s:design_concurrent_sched}). {\bf (3)} We implement and evaluate our system with real robots to demonstrate feasibility and gains (\S\ref{s:impl}, \S\ref{s:eval}).
\section{Background and Motivation}
\label{s:background}

This section provides a primer on physical AI, introduces the state-of-the-art foundation
models for physical AI and their shared properties
(\S\ref{s:physical_ai_models}), and then examines why existing serving
systems are suboptimal and the optimization opportunities that they leave
untapped (\S\ref{s:horizon} and \S\ref{s:exec_aware}).

Transformer-based foundation models have moved beyond the digital world~\cite{gpt4, gemini, claude, swe_agent, webarena} to transforming the physical world. Physical AI encompasses intelligent 
systems that perceive the real world and act on it. With a growing ecosystem of robotic
platforms, e.g., dexterous manipulator arms, mobile bases, and 
full-scale humanoids~\cite{figure, 1x, unitree}, physical AI is poised to shape many domains, including warehouses~\cite{amazon_robotics},
factories~\cite{tesla_optimus}, homes~\cite{1x, figure}, and healthcare~\cite{surgical_robot_survey}.

\subsection{Physical AI Models}
\label{s:physical_ai_models}

\begin{figure}[t]
  \centering
  \includegraphics[width=\linewidth]{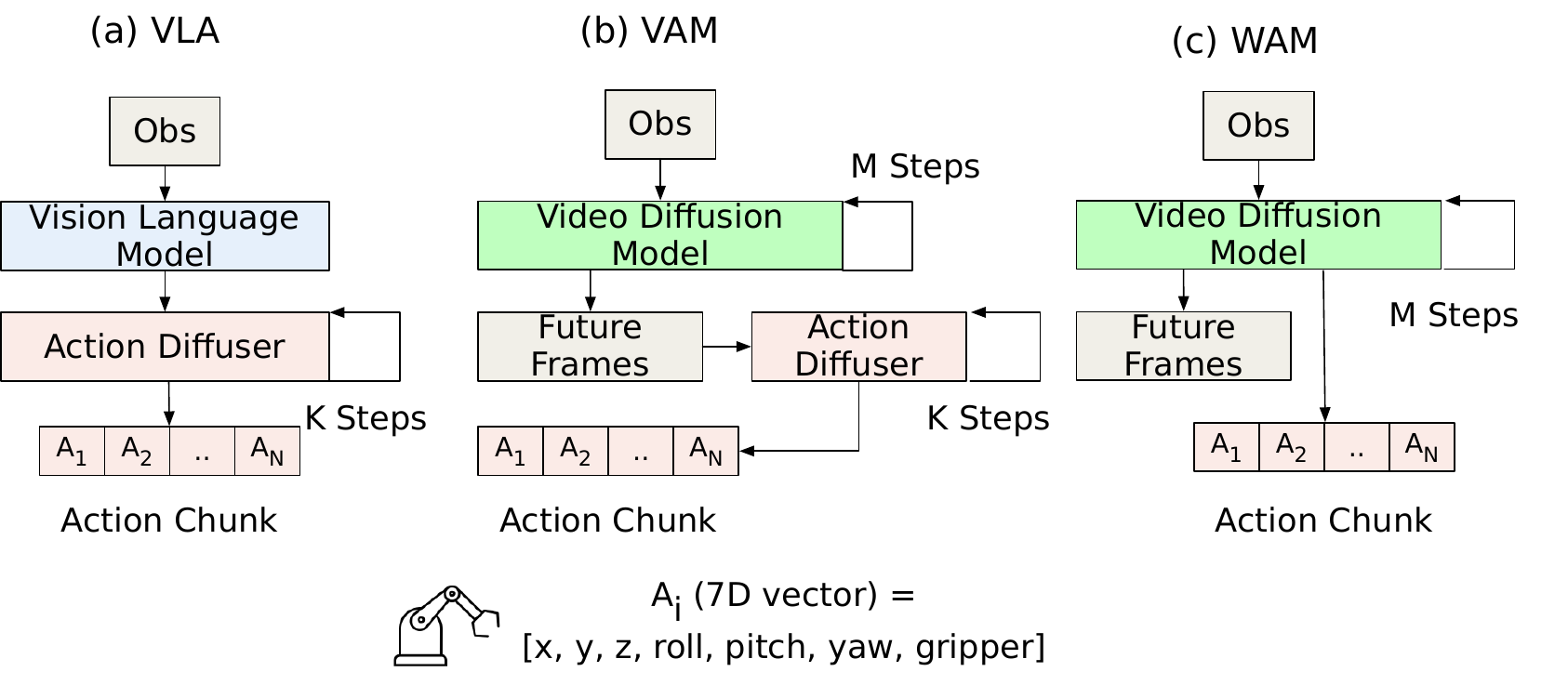}
  \caption{\small \bf The state of the art physical AI models.}
  \label{fig:physical_ai_models}
  \vspace{-10pt}
\end{figure}

Foundation models for physical AI map {\em observational inputs} to {\em actions}. Example observational inputs include camera images, robot states, and task instructions. Actions, on the other hand, consist of low-level
motor commands, such as gripper signals or joint positions, to advance the robot towards the instructed goal. 
Figure~\ref{fig:physical_ai_models} shows the family of physical AI models for robotic manipulation. 
\emph{Vision-Language-Action
models} (VLAs; ~\ref{fig:physical_ai_models}(a))~\cite{rt2, openvla, pi05, smolvla, groot} feed
observations through a pre-trained vision-language backbone and decode actions for the robot to execute  from its joint visual-semantic representation. \emph{Video Action Models}
(VAMs; ~\ref{fig:physical_ai_models}(b))~\cite{mimic-video,unified-VAM,S-VAM,DiT4DiT} replace the vision language backbone with a video diffusion model that first predicts future visual frames, then decodes actions based on the predicted frames. \emph{World Action Models}
(WAMs; ~\ref{fig:physical_ai_models}(c))~\cite{dream_zero,Fast-WAM,cosmos,gigaworldpolicy} uses a video diffusion model to generate future frames as well as actions simultaneously. Our work targets this family of frontier physical AI models (VLAs, VAMs, WAMs); we next explain their key properties and differences compared to LLMs used in digital AI.

\noindent{\bf 1) Action chunking.} Modern robots operate at control frequences of 30 Hz (or higher), thus making the typical paradigm of generating output tokens autoregressively one after the other, to be impractical for robotic AI. With robots demanding a new action every $\sim$33ms, and given the inference times of hundreds of millseconds, generating each action with a separate inference would cause the robot to pause after every action, leading to jerkiness and loss in utility. State-of-the-art physical AI models instead employ \emph{action chunking}: a single inference call produces a chunk of $N$ actions at the same time using diffusion.\footnote{We use ``diffusion'' broadly to include both denoising diffusion~\cite{diffusion_policy}, which iteratively removes noise from a corrupted sample, and flow matching~\cite{flow_matching}, which integrates a learned velocity field; both use a fixed number of steps to produce the full action chunk.} Its ability to enable high-frequency control of robots has made diffusion the dominant action-generating paradigm for all of the models in the physical AI family (Figure~\ref{fig:physical_ai_models}).

\begin{figure}[t]
  \centering
  \begin{subfigure}[t]{\linewidth}
    \centering
    \includegraphics[width=\linewidth]{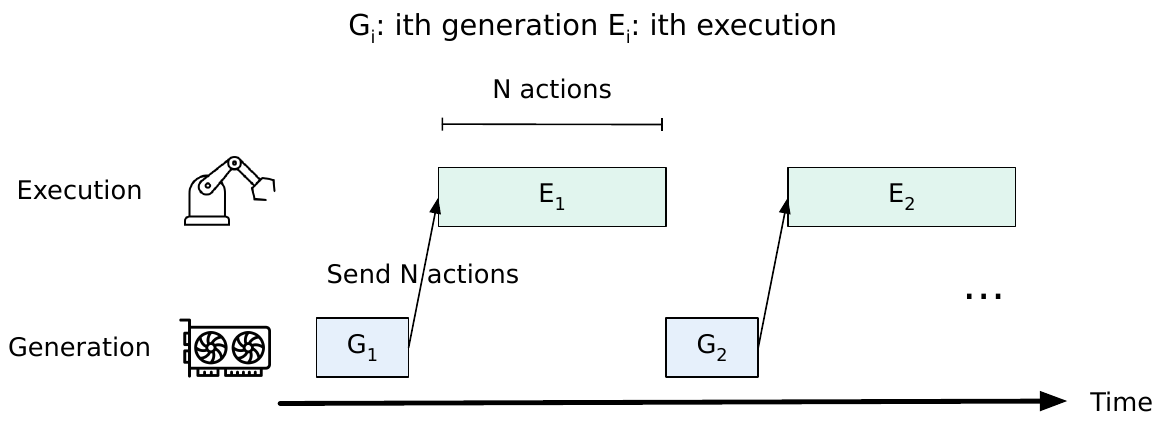}
    \caption{\small \bf Physical AI model serving is inherently multi-round: each round, the model generates a chunk of $N$ actions for robot execution. This generation--execution cycle repeats until the task completes.}
    \label{fig:exec_horizon}
  \end{subfigure}
  \vspace{0.5em}
  \begin{subfigure}[t]{\linewidth}
    \centering
    \includegraphics[width=0.9\linewidth]{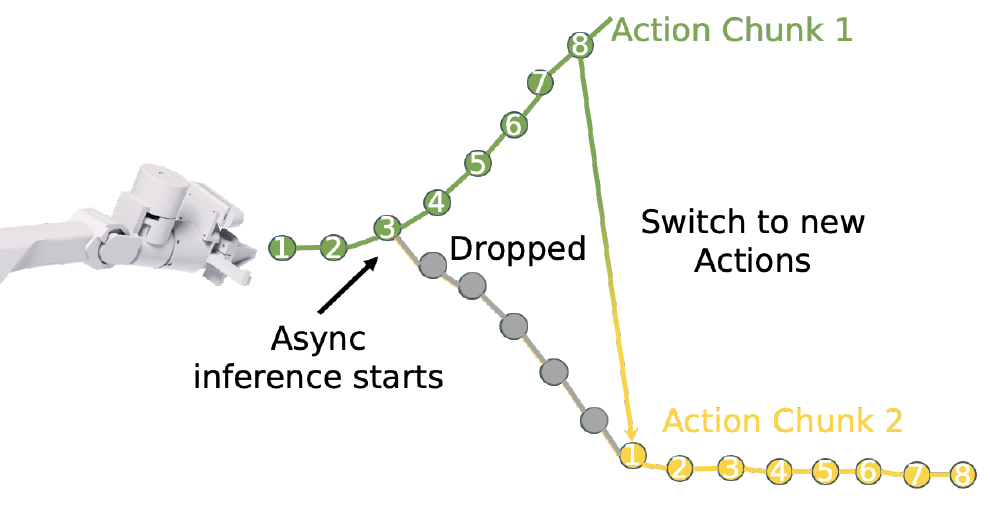}
    \caption{\small \bf Async inference overlaps the next inference with current execution, so a fresh action chunk is ready when the robot exhausts its actions. Actions in the overlap region (grey) are discarded to align with execution due to inference delay.}
    \label{fig:async_inference}
  \end{subfigure}
  \vspace{-12pt}
  \caption{\small \bf Physical AI model inference paradigm.}
  \label{fig:vla_inference}
  \vspace{-10pt}
\end{figure}

\noindent{\bf 2) Asynchronous multi-round inference.} A typical physical AI {\em task}, such as picking up a cup or setting a table cloth, requires multiple action chunks. With each chunk generated by an inference call, a physical AI task requires multiple {\em rounds} of inference in succession. A naive approach would generate a chunk of actions, and after the robot executes them, generate the next chunk with fresh observations (Figure~\ref{fig:exec_horizon}). Naturally, this leads to idling of the robot between inference rounds. To mitigate this, existing physical AI systems \cite{vlash, smolvla, rtc,dream_zero, gigaworldpolicy} apply \emph{asynchronous inference}, which launches the next inference while the robot is still executing the current chunk (Figure~\ref{fig:async_inference}), thereby overlapping action generation with action execution to avoid (or reduce) idling. 





\subsection{Execution Horizon}
\label{s:horizon}

As a direct consequence of the two points in (\S\ref{s:physical_ai_models}) -- action chunking and multi-round inference -- physical AI models suffer from {\em execution staleness}: all actions in a chunk are conditioned on a single observation captured at the start of each round, so later actions can become outdated and lead to degraded execution, e.g., collisions or missed grasping. This is unlike LLM based workloads in digital AI where generated tokens remain valid regardless of when they are consumed. To mitigate this staleness, robotic practitioners tune the {\em execution horizon, H}, which is a knob that controls how many actions are executed from each chunk. The set of actions in the chunk beyond the first $H$ are discarded.\footnote{Since the number of actions generated in a chunk is already baked into the model architecture at training time and set large for stable convergence~\cite{act, diffusion_policy}, tuning the execution horizon is a practically simple approach compared to retraining the model for a new action chunk size of $H$. 
} 

The execution horizon is a key parameter that allows {\em trading off accuracy with resource demand}. A shorter value of $H$ generates actions more frequently with fresher observations, thus improving accuracy but also proportionally increasing inference load. A longer $H$, on the other hand, lowers inference cost but risks executing stale actions for the robot, which impacts its accuracy for complex tasks (e.g., grasping under contact uncertainty) that require fresher actions.

\begin{figure}[t]
  \centering
  \includegraphics[width=0.85\columnwidth]{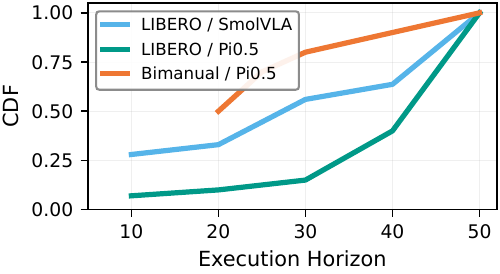}
  \vspace{-6pt}
  \caption{\small \bf The best per-task execution horizon varies widely across tasks and workloads. Each curve shows the CDF of the best per-task $H$ for a workload.}
  \label{fig:optimal_horizon}
\end{figure}

\begin{figure}[t]
  \centering
  \includegraphics[width=0.85\linewidth]{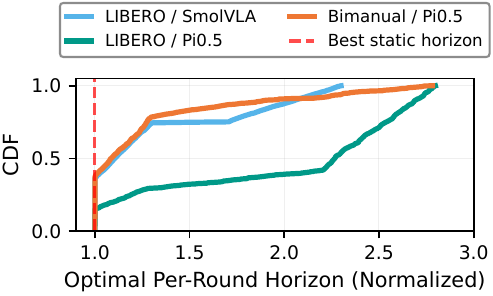}
    \vspace{-6pt}
  \caption{\small \bf The best per-task execution horizon varies within a task. CDF of the optimal per-round optimal execution horizon, normalized by the best per-task horizon.}
  \label{fig:horizon_cdf}
  \vspace{-10pt}
\end{figure}

\begin{figure*}[t]
  \centering
  \includegraphics[width=\linewidth]{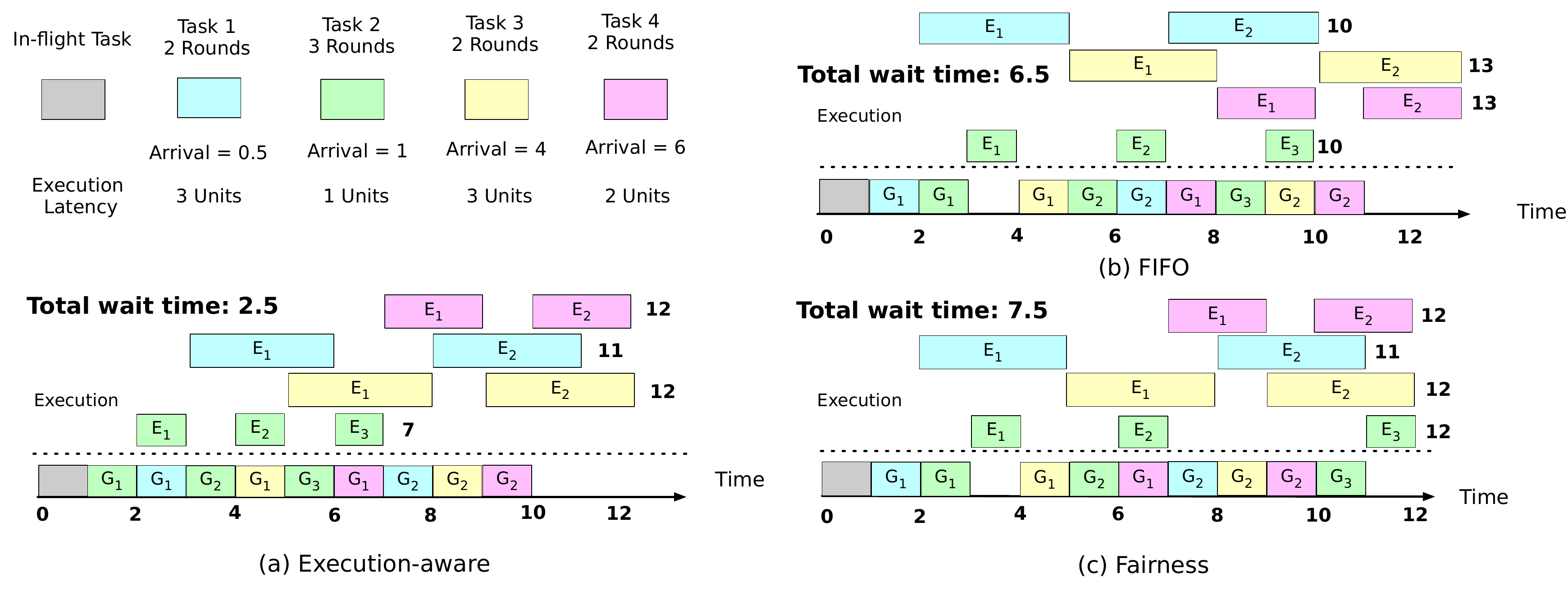}
    \vspace{-24pt}
 \caption{\small \bf We use four physical AI tasks to illustrate that execution-unaware scheduling leads to suboptimal
decisions. Task~2 (green) has longer total generation time but shorter
end-to-end latency. FIFO ignores task-level information and only
prioritizes request-level wait time, while Autellix~\cite{autellix}
misclassifies it as a long-running task and prioritizes it. Execution-aware scheduling correctly prioritizes Task~2, reducing total wait time to 2.5 units compared to 6.5 (FIFO) and 7.5 (Fairness).} 
  \label{fig:scheduling_example}
\end{figure*}

We quantify the variation across robotic tasks in the best value of $H$, i.e., the highest value of $H$ that achieves the highest accuracy. By scanning for the highest possible value of $H$, we also minimize the inference load. We use representative workloads from our evaluation (Table~\ref{tab:workloads}), including simulated and real bimanual robot tasks.  As Figure~\ref{fig:optimal_horizon} shows, $\sim40\%$ of tasks see no drop in accuracy even for $H \geq 40$ out of a maximum of 50. In the simulator with the Pi0.5 model, less than $10\%$ of tasks demand an execution horizon under 10. For the bimanual robot task, the minimum $H$ value needed is only 20 while $10\%$ of the tasks are fine with the horizon in excess of 40. The above results show not only a wide variation but also an opportunity to pick high values of $H$ (which will reduce inference load) without impacting accuracy for certain tasks.



Further, the best value of $H$ not only varies across tasks, but also {\em within the rounds of the same task}. The analysis above sets the same execution horizon across all the rounds of a task, which is reflective of the state-of-the-art practice in robotics deployments, for reasons of simplicity. Such a conservative value of $H$ typically reflects the most complex section (round) of a robotic task, but not all sections are equally complex. 
Intuitively, moving to a soda can is easier than the actual step of grasping the can (that requires the very short horizon), but moving to the can may take longer for a robot to execute because of the distance it has to traverse. This means that most physical AI serving deployments today run inference too frequently most of the time.


We quantify the magnitude of such a conservative choice and the opportunity for improvement in Figure \ref{fig:horizon_cdf}. 
We first collect the action traces using the task-optimal execution horizon set by Figure \ref{fig:optimal_horizon}. 
We then implement a simple offline algorithm that finds the round-optimal execution horizon for each round in the recorded traces. 
For each recorded round, we start with the task-optimal horizon. We increase it in small steps, we rerun the inference and measure the difference between the newly obtained trajectory and the pre-recorded trajectory of the robot (using cosine similarity). The round-optimal execution horizon is the largest execution horizon for which the cosine similarity remains above a threshold ($\geq 0.9$, tuned to preserve task accuracy).


Figure \ref{fig:horizon_cdf} plots the CDF of the ratios of the round-optimal versus task-optimal horizons for all rounds across all tasks. As shown, half the rounds can increase their horizon by $\geq 2\times$ while preserving accuracy. Even for the real bimanual setup, $20\%$ of rounds can increase $H$ by 1.5$\times-2.8\times$. These present a significant opportunity to achieve inference efficiency (by having less frequent inference rounds) without impacting accuracy, which existing serving systems do not exploit.

\subsection{Execution-aware scheduling}
\label{s:exec_aware}

As physical AI scales in the size of models and number of robots, the serving infrastructure for robot AI models is becoming a consolidated theater for multiplexing inference for multiple robots \cite{offload}. Building upon \S\ref{s:horizon} that discussed the resource optimization opportunities for a single robot's inference, we now discuss the scheduling opportunities in inference for multiple simultaneous physical AI models.

The latency of a physical AI task includes not only generation time (inference) but also execution, and as discussed, a task consists of multiple rounds of actions and inferences. Execution time can often be dominant depending on the execution horizon; at 30\,Hz, execution time is 300ms at $H{=}10$ but shoots to 1.6\,s  at $H{=}50$. This can shift a task from being generation-dominant to execution-dominant. However, existing serving frameworks are {\em unaware} of execution times. They schedule shared GPU resources in FIFO order of individual inference requests ~\cite{vllm, sglang, tensorrt-llm}, or for fairness (e.g., least
attained service) to equalize GPU time for generation among tasks and to mitigate starvation \cite{autellix}. This leads to sub-optimal latency of physical AI tasks.




Figure~\ref{fig:scheduling_example} presents an illustrative example of the above execution-unaware schedulers -- FIFO and fairness -- with a scheduler that is aware of execution times; the generation time is assumed to be the same for each inference. We ignore network latency and assume synchronous inference for simplicity. We consider four physical AI tasks, each with multiple rounds of generation (inference) and execution (actions), and each of the tasks has different execution times for its respective execution horizons, keeping with the observations in \S\ref{s:horizon}. Focusing on Task 2's latency (green), we can see the benefit of execution awareness. FIFO and fairness schedulers prioritize tasks with longer execution times ahead of Task 2, do not focus on the end-to-end task latency, and end up starving shorter tasks. Task 2's latency is 7 units with an execution-aware scheduler, compared to 10 and 12 with FIFO and fairness schedulers. Further, even the average task completion time is lower for the execution-aware scheduler. This is a result of the average time spent waiting by each robot to be only 0.6 units with the execution-aware scheduler, compared to 1.6 and 1.8 for the FIFO and fairness schedulers. Note that under synchronous inference, wait time equals queuing delay; we discuss the asynchronous case in \S\ref{s:design_concurrent_sched}.

\section{\name{}: System Design}
\label{s:design}

We design \name{}, a serving system tailored for physical AI workloads, with the guiding principle of being {\em execution-aware}: it optimizes the number of actions executed per round (execution horizon) to balance accuracy and efficiency, and incorporate execution-phase information into scheduling to minimize end-to-end task latency. Specifically, \name{} introduces two techniques: (1)~exposing model inference confidence as a fine-grained, per-round knob to dynamically select the execution horizon, achieving a better
accuracy-efficiency trade-off than any static horizon
(\S\ref{s:design_exec_horizon}), and (2)~an execution-aware scheduler
that leverages both generation and execution phase information to
minimize end-to-end latency across concurrent robotic requests
(\S\ref{s:design_concurrent_sched}).

\begin{figure}[t]
  \centering
  \includegraphics[width=\columnwidth]{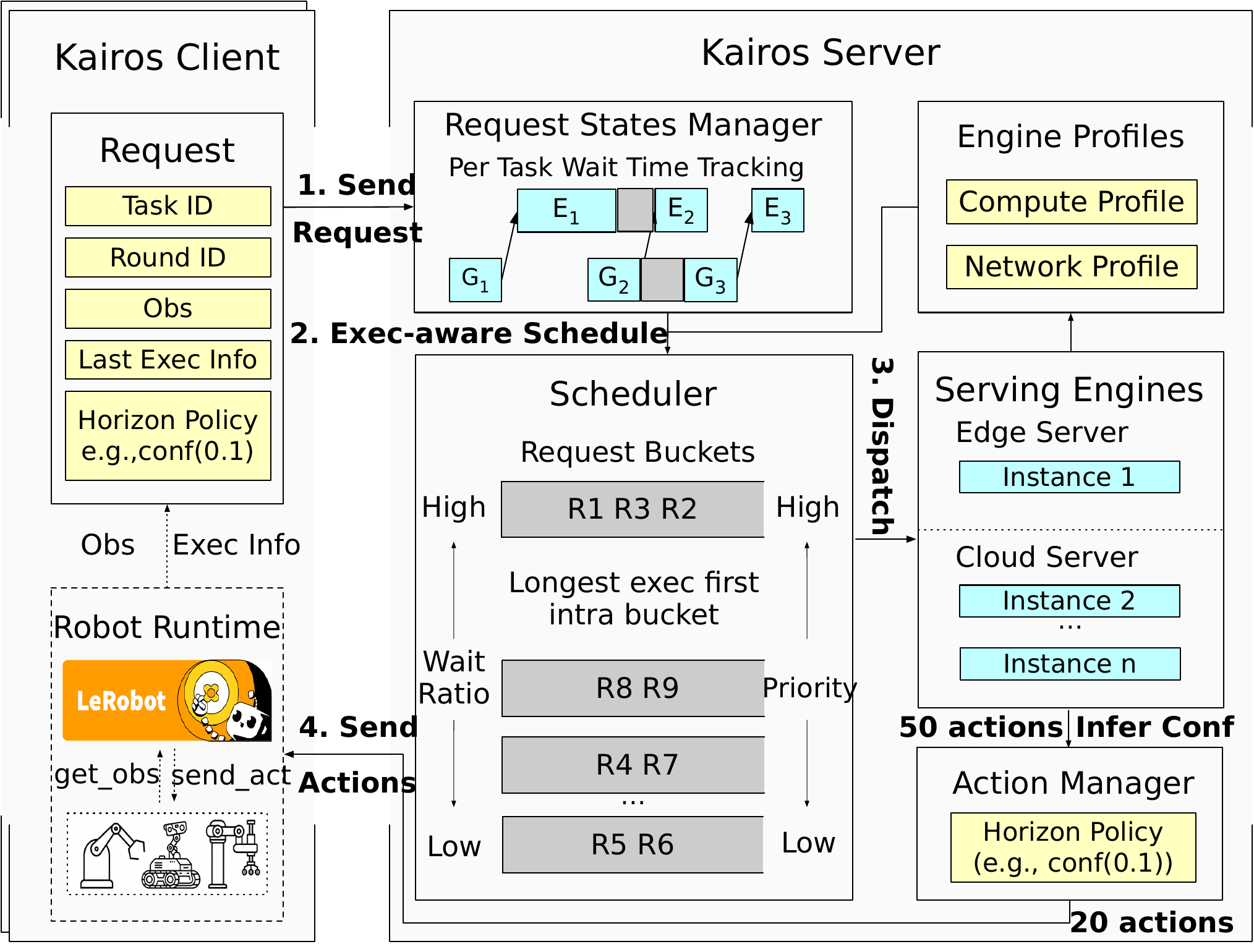}
  \vspace{-20pt}
  \caption{\small \bf System architecture.}
  \label{fig:sys_arch}
\end{figure}

Figure~\ref{fig:sys_arch} overviews \name{}'s architecture.
\name{} operates in a client-server model. Each client is co-located
with a robot runtime and communicates with the server in a
request-response loop: every round, the client sends a new observation
along with metadata about its previous execution phase (step 1 in Figure~\ref{fig:sys_arch}), and the server returns an action chunk (step 4) whose length is determined dynamically by \name{}'s execution horizon policy (\S\ref{s:design_exec_horizon}). \name{} by default exposes a diffusion confidence threshold as a user-configurable knob, but provides a flexible interface for users to configure their own execution horizon policy. On the server side, \name{} maintains each client task's lifecycle state (e.g., generation and execution timestamps of past rounds) to obtain per-task cumulative wait times, which feed into an execution-aware scheduler (step 2) that prioritizes requests for inference based on the ratio of wait times over its lifetime (considering both generation and execution (\S\ref{s:design_concurrent_sched}). The scheduler dispatches requests to a pool of serving engine instances (step 3) spanning edge and cloud tiers. We detail the client-server interface and server
components in \S\ref{s:impl}. 

\subsection{Dynamic Horizon with Inference Confidence}
\label{s:design_exec_horizon}


Recall from \S\ref{s:horizon} that the optimal execution horizon $H$ varies across tasks, and even across rounds within a single task, yet has to be identified against a reference trajectory that is unavailable at inference time. Prior work~\cite{diffdagger} approximates such a reference trajectory via self-consistency: generating multiple candidate action chunks to agree on one. However, this approach significantly inflates inference cost, often even undermining the efficiency gains that a dynamic execution horizon provides. In contrast, we make the following observation: the diffusion process of action generation itself provides a confidence signal for each action {\em at no extra cost}. 

Physical AI models generate the entire action chunk through $K$ iterative refinement steps. At each step, the model produces an update for \emph{every action} in the chunk. For example, with a 7-dimensional action space and a chunk size of $N{=}50$, each step produces a $50 \times 7$ matrix
of updates --- one 7D update vector per action. This process follows a well-established coarse-to-fine trajectory in which the early steps produce large and coarse corrections, and later steps contribute diminishing refinements. Prior work~\cite{adaptivediffusion, prophet} exploits this diminishing-update structure to accelerate inference by skipping
refinement steps for output regions (e.g., image patches) deemed confident once their values stabilize. Our key observation is
that, after $K$ steps, not every action in the chunk has
converged confidently: some actions exhibit diminishing updates throughout the refinement steps, while others still incur substantial updates at the final step. An action still being significantly revised at the last step is one
that the model is uncertain about. \name{} uses this as a per-action
confidence indicator --- not to skip denoising steps, but to
determine which actions in the generated chunk can be safely
executed and where to re-plan.

Concretely, after each generation round, \name{} exposes the per-action per-step diffusion updates $\mathbf{U}$ to an execution horizon policy module. The policy maps $\mathbf{U}$ to the execution horizon $H$ for the current round. This module is general: a static horizon policy simply ignores $\mathbf{U}$ and returns a fixed $H$, while dynamic policies can exploit $\mathbf{U}$ to adapt $H$ each round based on the confidence of the predicted actions. Any method that maps the intermediate diffusion updates or other intermediate inference states to the execution horizon decision can be plugged in, and this is configurable by the user, as shown in the horizon policy field in Figure~\ref{fig:sys_arch}. 



As a default instantiation of this module, \name{} provides a threshold-based policy that uses the diffusion updates $\mathbf{U}$ to examine each action's degree of convergence with a threshold $t$ as the tuning knob. Concretely, for each action in a generated chunk, we compare the magnitude of its final diffusion step update against the mean of its earlier updates. We use the final step as the reference because it is the update that directly produces the final action, making its magnitude a direct indicator of uncertainty. Walking sequentially from $A_1$, we stop at the first action $A_i$ whose final update exceeds $(1+t)$ times that mean and set the
execution horizon to $H{=}i{-}1$, retaining only the converged prefix $A_1, \ldots, A_{i-1}$. This is necessary because the robot
executes actions in order: an unconverged action at position $i$ compromises the trajectory from that point onward, regardless of
whether later actions appear individually converged.

\begin{figure}[t]
    \centering
    \includegraphics[width=0.9 \linewidth]{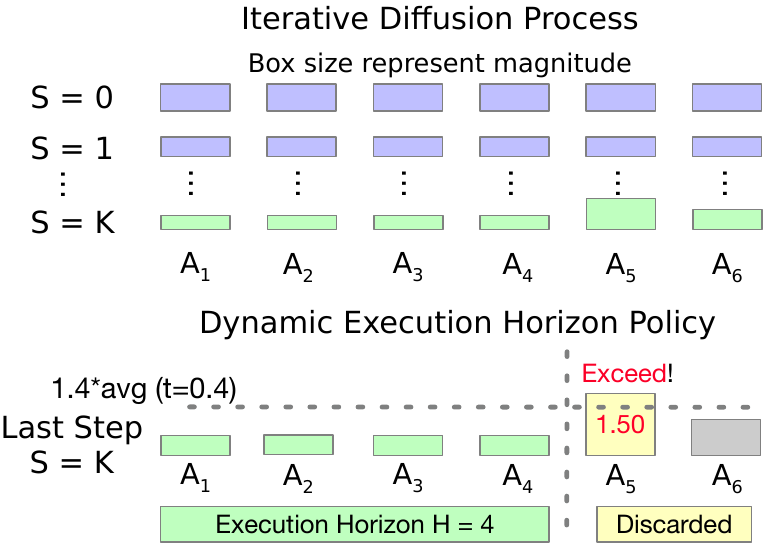}
   \caption{\small \textbf{Diffusion confidence and dynamic horizon
    selection. Box sizes show per-action update magnitudes at each step. The horizon policy scans from $A_1$, comparing each action's final update magnitude against $(1{+}t)$ times its mean over earlier steps. Here $A_5$ exceeds the threshold, setting $H{=}4$ and
    discarding $A_5$--$A_6$.} }
    \label{fig:confidence}
    \vspace{-10pt}
\end{figure}

As shown in Figure~\ref{fig:confidence}, actions $A_1$--$A_4$ exhibit small final updates, indicating convergence, while $A_5$ is the first whose final update is 50\% higher than its earlier mean, exceeding the $t{=}40\%$ threshold; so, \name{} retains $A_1$--$A_4$ and sets $H{ = }4$ for this round. \name{} sets $H = \max(H_{\text{thresh}}, H_{\min})$, where $H_{\min}$ is the smallest static horizon in the range we evaluate, ensuring a minimum number of actions per round. We observe that earlier actions are more likely to converge than later ones, with $H_{\text{thresh}} > H_{\min}$ in the majority of cases. We show in \S\ref{s:eval_confidence_knob} that using diffusion confidence as a tuning knob Pareto-dominates a static horizon knob across all our workloads.




\subsection{Execution-aware Scheduler}
\label{s:design_concurrent_sched}

To minimize end-to-end task latency when multiple physical AI tasks share a serving platform, the scheduler must reason about both generation and execution phases. Existing schedulers minimize wait time to achieve this, but define it solely in terms of queuing delays at the
generation phase: FIFO greedily minimizes per-request queuing delay, while fairness-oriented policies such as least attained service first~\cite{NUYENS2008286, 10.5555/3323234.3323274} and
Autellix~\cite{autellix} prioritize requests from tasks with lower accumulated generation time, reducing queuing delays for shorter tasks. However, these approaches ignore execution phases, which constitute a significant portion of end-to-end task latency and vary widely across tasks and across rounds (\S\ref{s:design_exec_horizon}). Moreover, with asynchronous generation and execution (\S\ref{s:physical_ai_models}), queuing delay at the generation side does not directly translate to added wait time for a task---if the robot is still busy executing, the queued generation request is not yet needed to be scheduled. The right metric is the delay relative to \emph{when the next action chunk is actually needed}. We first formalize wait time for physical AI tasks, then present \name{}'s scheduling policy.

\begin{figure}[t]
  \centering
  \includegraphics[width=0.9\columnwidth]{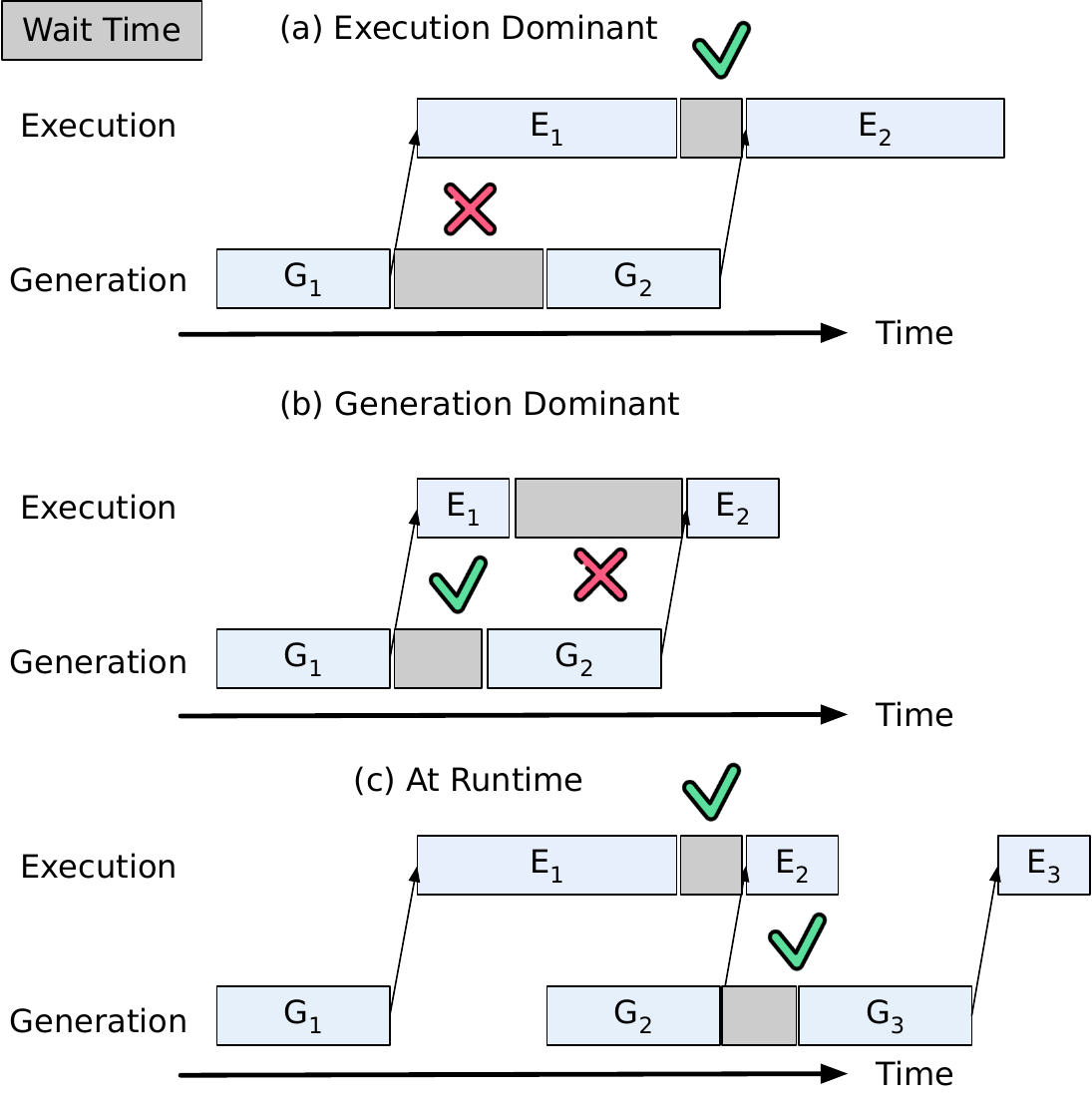}
  \vspace{-6pt}
  \caption{\small \bf Wait time measurement.
  (a)~Execution-dominated round: the optimal schedule overlaps $G_2$
  with ongoing $E_1$ fully while starting $G_2$ as late as possible, so $W$ between rounds 1 and 2 is the gap between $E_1$ ending and $E_2$ starting.
  (b)~Generation-dominated round: $W$ is the gap between $G_1$ ending and $G_2$ starting, since optimally $G_2$ begins right after $G_1$ with a fresh observation.
  (c)~\name{} identifies the dominant phase per round and measures
  $W$ accordingly.}
  \label{fig:wait_time}
  \vspace{-10pt}
\end{figure}

\subsubsection{Wait time of physical AI tasks.} To quantify the wait time of each task, we compare each task's actual progress against its optimal progress---the latency it would achieve by starting generation as late as possible while still maximally overlapping generation and execution. We treat each generation--execution round as a primitive: round~$i$ consists of a generation phase $G_i$ followed by an execution phase $E_i$, and the total wait time is the sum of the wait times between all consecutive rounds. For two consecutive rounds, the optimal timeline---and therefore where wait time is measured---differs depending on whether the first round is generation-dominated or execution-dominated. Consider the first round being execution-dominated (Figure~\ref{fig:wait_time}a): $E_1$ dominates $G_1$, and since diffusion inference latency remains constant, $E_1$ also dominates $G_2$. Therefore, the optimal schedule leverages async inference to start $G_2$ of the second round at the latest time possible, while fully overlapping with the ongoing $E_1$. The wait time between the two rounds is therefore the gap between $E_1$ ending and $E_2$ starting on the execution side. Recording the idle gap between $G_1$ and $G_2$ on the generation side would be incorrect: because execution dominates, a gap between $G_1$ and $G_2$ exists even under the optimal schedule.  Similarly, when the first round is generation-dominated (Figure~\ref{fig:wait_time}b), measuring the wait time on the execution side would be incorrect: even if $G_2$ of the second round starts as early as possible, a gap between $E_1$ and $E_2$ still exists because generation dominates. The earliest start time of $G_2$ possible could be right after $G_1$ with a fresh observation. The wait time is instead the gap between $G_1$ ending and $G_2$ starting on the generation side.  As Figure~\ref{fig:wait_time}c illustrates, \name{} therefore identifies the dominant phase each round and measures the wait time on that side.

To compute per-round wait times, \name{} maintains a per-task state record storing the sequence of
generation intervals $(G_0, G_1, \ldots)$ and execution intervals $(E_0, E_1, \ldots)$. With asynchronous generation, a task's round~$i$ generation request may be sent while round~$i{-}1$'s execution is still ongoing. The client piggybacks $E_{i-1}$'s start timestamp and the number of remaining actions on the generation request; since the robot executes actions at a fixed frequency, \name{} computes $E_{i-1}$'s end time from the request timestamp and the remaining action count. In this way, \name{} incrementally builds the complete generation and execution history of each task at runtime, with minimal bookkeeping overhead on both the client and server. From the per-round generation and execution history, \name{} computes each task's accumulated wait time and uses it to assign scheduling priorities, aiming to minimize wait time and therefore end-to-end latency.

\begin{algorithm}[t]
\DontPrintSemicolon
\caption{\small \bf Execution-aware scheduling}
\label{alg:scheduler}
\small
\KwIn{Pending inference requests $\mathcal{R}$ from physical AI tasks, buckets $B$, aging interval $A$, edge profile $\mathcal{P}_e$ with capacity $N_e$, cloud profile $\mathcal{P}_c$ with capacity $N_c$}
\KwResult{Dispatch assignments}
\BlankLine
\tcp{Phase 1: Compute wait ratio and bin requests}
\ForEach{$r \in \mathcal{R}$}{
    \ForEach{round $j$ with recorded $G_j, E_j$}{
        \leIf{$|G_j| \geq |E_j|$}{$W_j \gets G_{j+1}.\mathrm{start} - G_j.\mathrm{end}$}{$W_j \gets E_{j+1}.\mathrm{start} - E_j.\mathrm{end}$}
    }
    $w_r \gets \sum_j W_j / (t_{\mathrm{now}} - t_{\mathrm{start}}^r)$;~
    assign to $b_r \gets \lfloor w_r \cdot B \rfloor$\;
    \lIf{$r.\mathit{skipped} \geq A$}{assign to $b_r \gets \min(B-1, b_r + \lfloor r.\mathit{skipped} / A \rfloor)$}
}
\tcp{Phase 2: Within-bucket request ordering}
$\mathcal{S} \gets [\,]$\;
\For{$b = B\!-\!1$ \KwTo $0$}{
    $\hat{e}_r \gets |E^r_{\mathrm{last}}| \cdot (1 + r.\mathit{skipped})$ for each $r \in$ bucket $b$;~ sort by $\hat{e}_r$ desc; append to $\mathcal{S}$\;
}
\tcp{Phase 3: Shortest estimated delay guided placement}
$\mathcal{S}_e \gets \mathcal{S}[1:N_e]$;~ $\mathcal{S}_c \gets [\,]$\;\label{alg:line:placement_start}
\ForEach{$r \in \mathcal{S}[N_e\!+\!1:]$}{
    \lIf{cloud latency $< $ edge delay \textbf{and} $|\mathcal{S}_c| < N_c$}{append $r$ to $\mathcal{S}_c$}
}\label{alg:line:placement_end}
\lForEach{$r \in \mathcal{S}_e \cup \mathcal{S}_c$ with stale obs}{$r.\mathit{obs} \gets \texttt{fetch\_obs}(r.\mathit{client})$}
Dispatch $\mathcal{S}_e$ to edge, $\mathcal{S}_c$ to cloud\;
\ForEach{$r \in \mathcal{R}$}{
    \leIf{$r \in \mathcal{S}_e \cup \mathcal{S}_c$}{$r.\mathit{skipped} \gets 0$}{$r.\mathit{skipped} \gets r.\mathit{skipped} + 1$}
}
\end{algorithm}

\subsubsection{Scheduling policy.} Given per-round wait times, the schedule decides how to prioritize pending generation requests with the goal of minimizing wait times across concurrent tasks. Intuitively, given per-task wait times, a natural approach is to prioritize by accumulated wait (similar to least attached service), but this biases toward long-running tasks that naturally accumulate more wait, preventing shorter ones from being scheduled. Instead, we borrow insights from Highest Response Ratio Next~\cite{os-principle}, and normalize wait time by each task's elapsed lifetime to prioritize high \emph{wait ratio} requests to minimize wait time while improving fairness. 
\begin{equation}
  wr \;=\; \frac{\sum_{j} W_j}{\,t_{\mathrm{now}} - t_{\mathrm{start}}\,}
  \label{eq:wait_ratio}
\end{equation}
where $W_j$ is the wait time for round~$j$.  A high wait ratio indicates that a task has been disproportionately delayed and should be prioritized; a low one indicates the client has been well-served relative to its lifetime. Algorithm~\ref{alg:scheduler} summarizes the full scheduling policy; we describe the details below. 

\begin{figure}[t]
  \centering
  \includegraphics[width=\columnwidth]{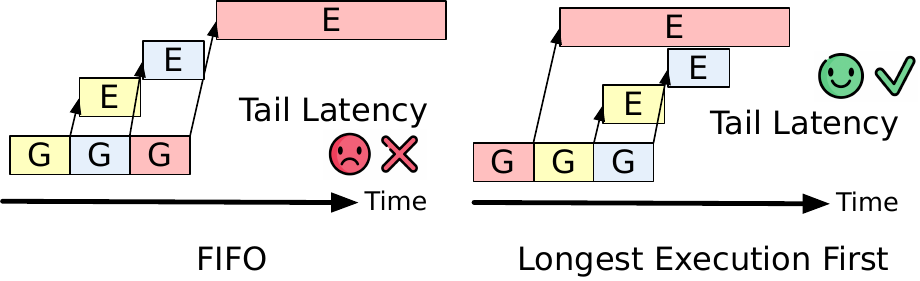}
  \vspace{-21pt}
  \caption{\small\bf Within-bucket ordering by estimated execution latency. \textbf{Left:} FIFO ordering can leave long-execution tasks to the end. \textbf{Right:} longest execution first reduces the tail latency while keeping the average latency.}
  \label{fig:intra_bin}
  \vspace{-5pt}
\end{figure}

\begin{figure}[t]
  \centering
  \includegraphics[width=0.88\linewidth]{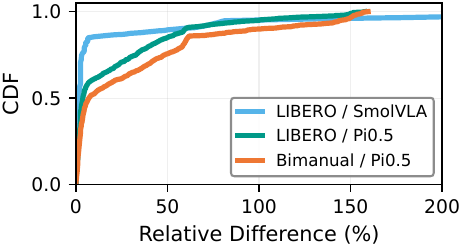}
  \vspace{-8pt}
  \caption{\small \bf CDF of the relative difference between consecutive-round execution latencies. Across workloads, 52.3--85.4\% of rounds have less than $10$\% deviation.}
  \label{fig:exec_latency_corr}
  \vspace{-15pt}
\end{figure}

Scheduling by continuous wait-ratio values can degrade into round-robin when many tasks have similar ratios~\cite{autellix}. The scheduler, therefore, uses a two-level hierarchical priority (Algorithm~\ref{alg:scheduler}).  The scheduler uses a two-level hierarchical priority. At the coarse level, it discretizes the $[0,1]$ wait-ratio range into $B$ (default 10) equal-width buckets, bins each generation request by its task wait ratio, and processes buckets from highest to lowest. Within each bucket, ties are broken by estimated execution latency (descending). The reason is that scheduling execution-dominant tasks first prevents long execution phases from starving, which can reduce tail latency while maintaining the average latency  (Figure~\ref{fig:intra_bin}, right vs.\ left). However, this ordering requires knowing the execution duration before actions are generated. To estimate the current round's execution latency, our insight is that physical execution is continuous---a robot's motion phase changes gradually across rounds---so the last observed execution duration is a reliable predictor of the current round.  Figure~\ref{fig:exec_latency_corr} validates this: 52.3--85.4\% of consecutive rounds differ by less than 10\%; only 4.9--10.1\% of rounds exhibit a relative difference exceeding 100\%, which corresponds to the cases with sharp transition boundaries where they shift between execution and generation dominant. 


To prevent starvation, the scheduler tracks the number of consecutive rounds in which a task's generation request is not selected; after every $A$ skipped rounds, the request is promoted by one bucket, guaranteeing progress regardless of its current wait ratio (\hyperref[alg:scheduler]{line~7}). However, bucket promotion alone is insufficient for generation-dominant tasks: even in the highest-priority bucket, a short execution duration would always lose the within-bucket ordering. The scheduler thus scales the estimated execution latency by an aging factor $(1 + \mathit{skipped})$, preventing requests from generation-dominant tasks to be indefinitely starved (\hyperref[alg:scheduler]{lines~8--10}). Once a task's generation is scheduled, its skip counter is reset to 0 (\hyperref[alg:scheduler]{line~19}).

Finally, before dispatching to the serving engine, if a request's observation has grown stale due to queuing, the scheduler re-fetches it from the client to ensure an up-to-date observation to incorporate any observation changes during its queuing delay(\hyperref[alg:scheduler]{line~16}). The scheduler then batches dispatched requests up to the maximum batch size, determined by the saturation point of the latency---batch size profile ($\mathcal{P}_e$ or $\mathcal{P}_c$) beyond which additional batching increases latency without improving throughput. Once dispatched, a request's skip counter is reset to zero (\hyperref[alg:scheduler]{line~19}). 

\subsubsection{Support hybrid edge and cloud serving.}
Physical AI tasks prefer co-located edge servers to avoid
excessive network latency to the cloud~\cite{fastirunvla}.
However, edge capacity is inherently limited: as fleet size grows
or bursty workloads spike demand, edge queues deepen, and latency
degrades. If done carefully, a hybrid edge--cloud deployment can
absorb these overflow bursts on cloud GPUs, trading network
round-trip cost for shorter queuing delay and faster computing. Many physical AI models are stateless---conditioning only on the current observation and
task description---and provide a more flexible placement
opportunity compared with LLM serving, so requests can be freely
placed on any available instance. Unlike autoregressive LLM inference where output length varies unpredictably, physical ai models run a fixed number of diffusion iterations, enabling \name{} to profile latency-vs-batch-size curves
($\mathcal{P}_e$, $\mathcal{P}_c$) offline for each serving tier.
The scheduler prefers to dispatch to the edge server and
offloads to the cloud only when the estimated cloud
latency---including network round-trip---is lower than the
expected queuing delay at the edge (\hyperref[alg:scheduler]{lines~\ref*{alg:line:placement_start}--\ref*{alg:line:placement_end}}).
For models that maintain history across rounds~\cite{dream_zero, memoryvla, cosmos}, this estimation additionally includes historical data transfer cost.


\section{Implementation}
\label{s:impl}


To our knowledge, no existing serving system supports online serving of concurrent physical AI tasks. We reimplement state-of-the-art scheduling techniques from LLM and multi-round agent serving systems as baselines (\S\ref{s:eval_setup}). We build \name{} in 3.7K lines of Python atop LeRobot~\cite{lerobot}, a popular open-source framework for physical AI, reusing its abstractions for simulated and real robots. We next describe the client-server interface in \name{}'s design (\S\ref{s:design}).

\para{Client.}
The \name{} client integrates with LeRobot's robot runtime through a lightweight shim that exposes two functions: \texttt{get\_obs} for capturing the current observation and execution information, and \texttt{send\_act} for streaming actions to the robot controller. The client communicates with the server via gRPC, attaching a unique
\emph{Task~ID}, a monotonically increasing \emph{Round~ID}, the current
observation, and \emph{Last Exec Info} reporting the start and end timestamps of the previous execution phase to each request. \name{} uses the diffusion-confidence-based execution horizon policy (\S\ref{s:design_exec_horizon}) by default, but exposes a flexible \emph{Horizon Policy} field for user customization. The Last Exec Info,
Round~ID, and Horizon Policy fields are what distinguish \name{}'s interface from a standard physical-AI inference-serving API such as LeRobot's: they grant the server continuous visibility into the physical-world phase of each request's lifecycle, enabling the execution-aware scheduling described in \S\ref{s:design_concurrent_sched}.

\para{Server.}
The server is built as an asynchronous event loop. Models are compiled with
\texttt{torch.compile} to optimize inference latency. The
\emph{Execution-aware Scheduler} assigns priorities and dispatches
requests to engine instances based on \emph{Engine Profiles}, which
record per-instance compute (batch size vs.\ latency) and network
latencies from offline profiling. Dispatched requests are
dynamically batched (up to the engine's specified max batch size) to maximize throughput. The \emph{Action Manager} sits on the return path: after an engine produces an action chunk, it applies the client's horizon policy and returns the trimmed action sequence to the client.
\section{Evaluation}
\label{s:eval}

We evaluate \name{} across a wide range of physical AI workloads, models, varying loads, and different deployment setups. Our key findings are: 
\begin{itemize}[leftmargin=*]
\item \textbf{Diffusion confidence Pareto-dominates static horizon as an efficiency--accuracy knob}, yielding up to 2.67$\times$ longer execution horizons at matched accuracy or up to 30\% accuracy gains at matched horizons (\S\ref{s:eval_confidence_knob}).
\item \textbf{\name{} scales to increasing online serving loads}, reducing average latency by 31.8--66.5\% over both baselines at peak load (up to 88.4\% P25, 52.0\% P95) (\S\ref{s:eval_sched_e2e}).
\item \textbf{\name{} scales with robot fleet size}, with growing average latency reductions of 20.4\% (10 robots) to 42.8\% (100 robots) 
\item \textbf{\name{} generalizes to diverse serving setups}, reducing average latency by 36.9--47.7\% over edge-only and 51.9--67.9\% over cloud-only in a hybrid edge--cloud setup (\S\ref{s:eval_sched_e2e}).
\end{itemize}

\subsection{Experimental Setup}
\label{s:eval_setup}

\para{Physical AI workloads.}
For simulation workloads, we use three common benchmarks:
(1)~\textbf{LIBERO}~\cite{libero}, a suite of 40 language-conditioned manipulation tasks across four categories: spatial reasoning, object manipulation, procedural knowledge and long-horizon planning on a simulated Franka Panda~\cite{franka} arm;
(2)~\textbf{Meta-World}~\cite{metaworld}, 50 manipulation tasks on a simulated Sawyer~\cite{sawyer} arm divided in easy, medium and hard difficulty categories; and
(3)~\textbf{NVIDIA Isaac Lab}~\cite{isaaclab}, a GPU-accelerated robot
learning framework built on Isaac Sim~\cite{isaacsim} for high-fidelity
robotic simulation, where we evaluate on a simulated Fourier
GR1~\cite{gr1} humanoid robot across 32 manipulation tasks spanning two categories in different environments. (4)~\textbf{RoboTwin 2.0}~\cite{robotwin}, a scalable bimanual manipulation benchmark with structured domain randomization across 50 dual-arm tasks spanning multiple robot embodiments; and
(5)~\textbf{SIMPLER}~\cite{simpler}, a real-to-sim evaluation suite that mirrors common real-robot setups and exhibits strong correlation with real-world policy performance. For real-robot experiments, we deploy on a customized \textbf{bimanual SO-101}~\cite{so101} setup, where one arm picks up an object and hands it to the other arm, which then places it into a box. We construct tasks by repeating this handoff-and-place procedure 20 times with varied initial object placements. 

\para{Models.} We pair these workloads with 6 models---four VLAs (Pi0.5~\cite{pi05}, SmolVLA~\cite{smolvla}, GR00T N1.5~\cite{groot}, XVLA~\cite{xvla}), one WAM (Fast-WAM~\cite{Fast-WAM}), and one VAM (mimic-video~\cite{mimic-video})---to evaluate \name{} on architecturally distinct model families. Table~\ref{tab:workloads} summarizes the workload model pairs.

\para{Testbed.}
All serving experiments run on an edge server equipped with an NVIDIA RTX A6000 GPU; edge--cloud experiments use an NVIDIA A100 (80GB) GPU in the cloud server. Following existing practice in physical AI serving~\cite{fastirunvla,lerobot}, we set the robot to the edge server gateway via Wi-Fi~7 (2.50\,ms base latency, 2\,Gbps upload, 3\,Gbps download) and the edge gateway to the cloud server via a WAN link (100\,ms base latency, 1\,Gbps symmetric). We evaluate additional edge--cloud network configurations in \S\ref{s:microbenchmarks}.

\para{Baselines and metrics.}
We compare \name{} against two baselines:
(1)~FIFO, a first-come-first-served scheduler and
(2)~Autellix~\cite{autellix}, a multi-round-aware scheduler that prioritizes requests with lower accumulated generation time. Both are assigned the largest static execution horizon $H$
that maintains the highest accuracy, tuned offline, and held constant throughout a task. Both baselines represent the state of the practice: FIFO is the default policy in existing LLM serving frameworks~\cite{vllm, sglang, tensorrt-llm}, while Autellix is the state-of-the-art scheduler for multi-round agent workloads but is unaware of the physical execution phase. For evaluation, we report two metrics: \emph{accuracy}, the fraction of tasks completed successfully, and \emph{end-to-end latency}, the wall-clock time from the first inference request to the last action executed, reported at the average, P25, and P95 over all tasks.

\para{Evaluation Methodology.} We evaluate \name{} with real and simulated robots. However, our serving scalability evaluation requires varying the number of concurrent robots, but co-locating many simulator instances on a single machine distorts execution timing and task accuracy, and physical robots are expensive to provision at scale. We therefore adopt a trace-driven evaluation. Specifically, we first execute each task in isolation---including on a physical robot and via a single simulator instance---and record a trace: the complete observation--inference--execution sequence along with the action index at which each inference request is triggered. At evaluation time, we replay many such traces concurrently against the serving system; each replayed application issues its inference request only at the recorded action index, faithfully preserving the original execution horizon. As in prior work~\cite{sglang,vllm}, we sample from collected task traces and simulate task arrivals using a Poisson process with varying arrival rates to capture different load conditions. 

\begin{table}[t]
  \centering
  \small
  \caption{\small \bf Summary of model workload pairs.}
  \vspace{-10pt}
  \label{tab:workloads}
  \setlength{\tabcolsep}{5pt}
  \begin{tabular}{ll}
    \toprule
    \textbf{Robotic Benchmarks} & \textbf{VLA Model(s)} \\
    \midrule
    LIBERO / simulator        & Pi0.5, SmolVLA, XVLA \\
    MetaWorld / simulator           & SmolVLA \\
    Isaac Lab / simulator      & GR00T N1.5 \\
    Real robot / Bimanual SO101 (real)       & Pi0.5 \\
    RoboTwin / simulator  & Fast-WAM \\
    SIMPLER / simulator &  mimic-video \\
    \bottomrule
  \end{tabular}
  \vspace{-10pt}
\end{table}


\subsection{Effectiveness of Diffusion Confidence}
\label{s:eval_confidence_knob}

\begin{figure}[t]
    \centering
    \includegraphics[width=\linewidth]{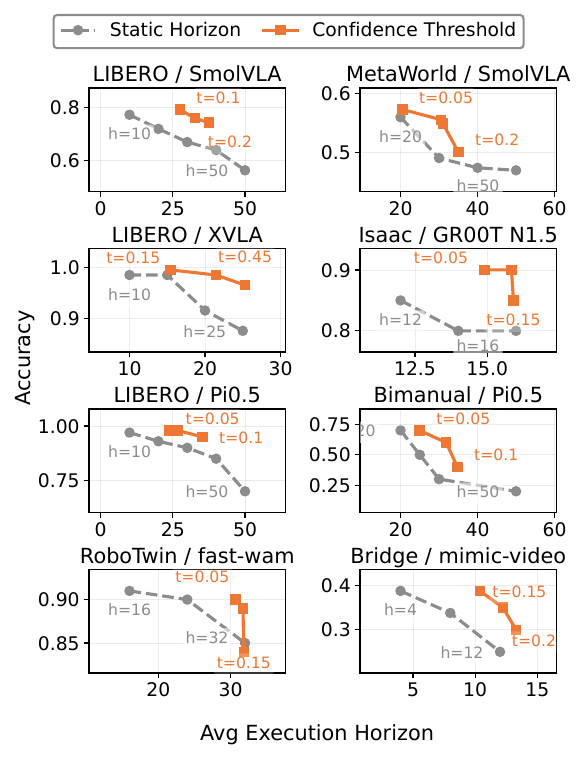}
    \vspace{-20pt}
    \caption{\small \bf Accuracy--efficiency trade-off across six workloads
  (Table~\ref{tab:workloads}). Each gray point corresponds to a
  fixed horizon $h$; each orange point corresponds to a confidence
  threshold $t$. The confidence-based policy consistently
  Pareto-dominates the static horizon across all workloads.}
    \label{fig:tradeoff}
    \vspace{-15pt}
\end{figure}

For each workload, we sweep the static horizon $h$ and the confidence threshold $t$, running 10 trials per configuration to account for randomness in simulation and the diffusion process, and plot task
accuracy against the average execution horizon---a proxy for system efficiency, since shorter horizons yield more frequent inference. Figure~\ref{fig:tradeoff} shows the results across all six workloads. The fine-grained confidence knob (orange) consistently Pareto-dominates the static execution horizon (gray). At comparable accuracy, it achieves up to 2.67$\times$ longer average execution horizons; at comparable horizons, it improves accuracy by up to 30\%. The gap demonstrates the benefit of a fine-grained, per-round execution horizon over a static, per-task one.

\subsection{End-to-End System Performance}
\label{s:eval_sched_e2e}

In this section, we evaluate \name{}'s end-to-end serving performance under two deployment scenarios: (1)~an online setting where tasks arrive according to a Poisson process at varying rates, evaluated on both an edge-only server and an edge--cloud configuration; and (2)~an offline setting where a fixed set of tasks is executed back-to-back by robot fleets of varying sizes. To focus on evaluating serving efficiency, we configure the execution horizon policies of \name and baselines using the accuracy--efficiency trade-off results from \S\ref{s:eval_confidence_knob}: for the static-horizon baselines (FIFO and Autellix), we assign each workload the largest fixed execution horizon $H$ that achieves the highest accuracy for each task; for \name{}, we select the highest confidence threshold $t$ that matches or exceeds that accuracy. Across all workloads, \name{} consistently achieves lower latency than both baselines, with the gap widening at higher task rates. 

\begin{figure*}[t]
    \centering
    \includegraphics[width=\linewidth]{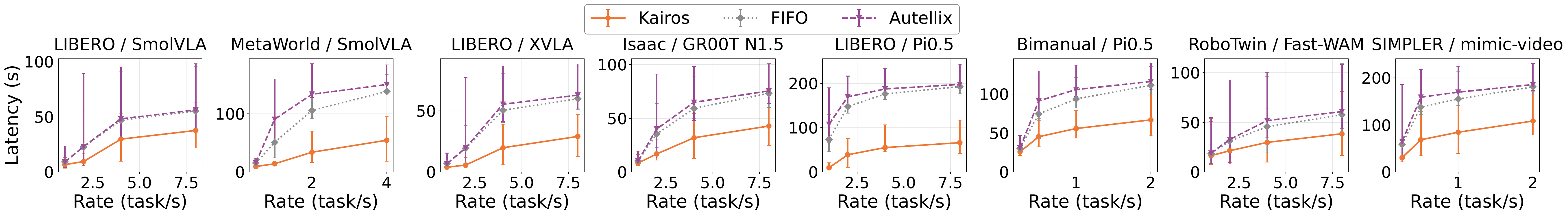}
        \vspace{-20pt}
    \caption{\small \bf Average end-to-end latency under increasing task arrival rates across workloads. Error bars show P25 and P95.}
    \label{fig:aggregate_rate}
    \vspace{0pt}
\end{figure*}

\begin{figure*}[t]
    \centering
    \includegraphics[width=\linewidth]{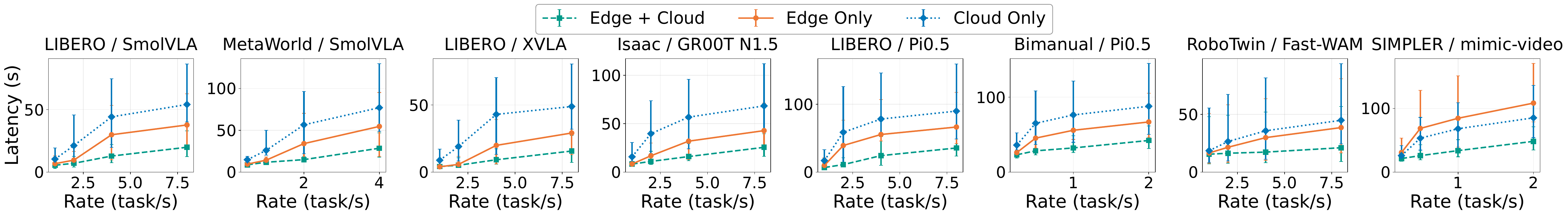}
    \vspace{-20pt}
    \caption{\small \bf Effect of edge--cloud offloading on average end-to-end latency for increasing arrival rates. Error bars show P25-90.}
    \label{fig:aggregated_offload}
\end{figure*}

\para{Online serving with varying request rates.} 
Figure~\ref{fig:aggregate_rate} shows the end-to-end latency performance under varying system loads across physical AI workloads with a single edge server. For each workload, we sweep the task arrival rates from low to high and report the average latency, with error bars spanning the P25 and P95 range. Across all workloads, \name{} consistently achieves lower latency than both baselines, with the gap widening as task arrival rates scale. Notably, Autellix consistently performs worse than FIFO, as it
assigns priority solely based on total generation time, which can starve the very short tasks it is designed to prioritize. Under the highest load, \name{} reduces average latency by 31.8--66.5\%, P25 latency by 39.5--88.4\%, and P95 latency by 22.2--52.0\% over both baselines. Table~\ref{tab:benchmarks-latency-reduction} reports the full latency reduction results at the highest arrival rate across all workloads. We further ablate two key components of \name{}: the dynamic execution horizon and the execution-aware scheduler. 

First, replacing the \name{} scheduler with either baseline scheduler while keeping the dynamic execution horizon alone reduces P25 latency by 3.9--58.0\%, average latency by 15.1--54.9\%, and P95 latency by 18.9--49.8\% over its static-horizon counterpart. This confirms that the dynamic horizon directly reduces total serving load by reducing inference requests. However, a latency gap up to 43.9\% P25, 22.3\% average, and 12.2\% P95 remains compared to the full \name{} system, showing that dynamic execution horizon alone can not compensate for execution-unaware scheduling.

Second, removing the dynamic execution horizon, \name{} still yields substantial reductions in P25 (21.1--63.5\%), average (10.9--39.3\%), and P95 (4.1--13.2\%) latency over both static baselines, validating the effectiveness of execution-aware scheduling. Notably, \name{} (static) achieves higher P25 reductions than the baseline schedulers paired with the dynamic horizon (up to 26.9\%) in the majority of workloads. This validates that \name{}'s execution-aware scheduler, with visibility into both generation and execution phases, correctly identifies and prioritizes the truly short requests--- without coming at the expense of longer tasks. Finally, \name{} (which combines both components) delivers the largest latency reductions.

\para{Online serving across hybrid edge-cloud deployments.}
Figure~\ref{fig:aggregated_offload} further evaluates \name{} under hybrid edge--cloud deployments, where \name{} dynamically offloads requests to a cloud GPU (A100 80GB) over a 100\,ms WAN link (\S\ref{s:eval_setup}). At the highest arrival rate, \name{} (hybrid) reduces average latency by 36.9--47.7\% over \name{} (edge-only) and 51.9--67.9\% over \name{} (cloud-only) across all workloads, similarly for P25 and P95. Cloud-only serving consistently performs the worst due to the WAN round-trip overhead for every request, with the exception of mimic-video, where the cloud GPU's faster compute outweighs the WAN cost and makes cloud-only more efficient than edge-only.

These results show that \name{}'s scheduler generalizes to heterogeneous serving settings, maintaining each task's generation and execution bookkeeping regardless of each inference request's placement and inference latency, effectively using extra cloud capacity to reduce latency.

\begin{figure*}[t]
    \centering
    \includegraphics[width=\linewidth]{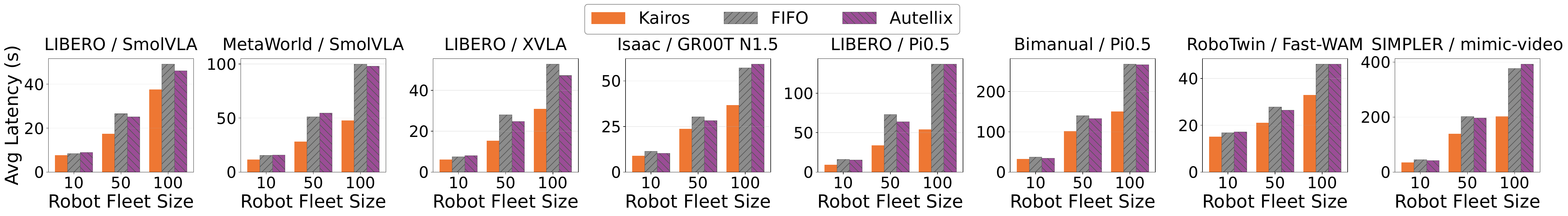}
        \vspace{-20pt}
    \caption{\small \bf Average end-to-end latency as the dedicated robot fleet scales from 10 to 100 concurrent robots.}
    \label{fig:aggregate_scale}
    \vspace{-8pt}
\end{figure*}

\para{Offline serving with varying fleet sizes.} In addition to evaluating \name{} as a shared online serving platform, we evaluate it in dedicated offline deployments, where a fleet of $n$ robots collectively executes a fixed set of tasks back-to-back. We vary $n$ from 10 to 100 robots to measure \name{}'s performance across different deployment scales. Figure~\ref{fig:aggregate_scale} shows the results. \name{} consistently outperforms both baselines across all fleet sizes and workloads, and the gap widens as the fleet scales up: at 10 robots, \name{} reduces average latency by 20.4--21.4\% over two baselines; at 50 robots, the reduction grows to 33.9--37.8\%; and at 100 robots, it reaches 41.0--42.8\%. This widening gap demonstrates that \name{} scales effectively with fleet size, as the dynamic execution horizon reduces the total number of inference requests while the execution-aware scheduler more effectively utilizes the serving budget under increasing contention.


\begin{table}[t]
\caption{\small \bf Latency reduction over static baselines at the highest arrival rate. Results are for \name{} and ablations removing the scheduler or dynamic execution horizon.}
\vspace{-4pt}
\label{tab:benchmarks-latency-reduction}
\centering
\scriptsize
\setlength{\tabcolsep}{2pt}
\renewcommand{\arraystretch}{1.15}

\begin{subtable}[t]{\columnwidth}
\subcaption{\small vs.\ FIFO}
\label{tab:latency-reduction-fifo}
\centering
\begin{tabular}{l >{\columncolor{p25color}}c >{\columncolor{avgcolor}}c >{\columncolor{p95color}}c >{\columncolor{p25color}}c >{\columncolor{avgcolor}}c >{\columncolor{p95color}}c >{\columncolor{p25color}}c >{\columncolor{avgcolor}}c >{\columncolor{p95color}}c}
\toprule
& \multicolumn{3}{c}{\name{}} & \multicolumn{3}{c}{\makecell{\name{} w/o \\ dynamic horizon}} & \multicolumn{3}{c}{\makecell{\name{} w/o \\ scheduler}} \\
\cmidrule(lr){2-4}\cmidrule(lr){5-7}\cmidrule(lr){8-10}
Workload
& \tiny P25 & \tiny Avg & \tiny P95
& \tiny P25 & \tiny Avg & \tiny P95
& \tiny P25 & \tiny Avg & \tiny P95 \\
\midrule
LIBERO/SmolVLA
& 39.5\% & 31.8\% & 34.6\%
& 21.1\% & 10.9\% &  9.9\%
&  3.9\% & 16.8\% & 29.0\% \\
LIBERO/XVLA
& 74.5\% & 51.4\% & 45.2\%
& 39.4\% & 17.8\% &  4.5\%
& 44.6\% & 41.9\% & 37.5\% \\
LIBERO/Pi0.5
& 76.4\% & 65.6\% & 51.9\%
& 57.0\% & 33.0\% &  5.7\%
& 58.0\% & 54.9\% & 49.6\% \\
MetaWorld/SmolVLA
& 86.9\% & 60.5\% & 43.0\%
& 59.0\% & 34.1\% &  4.3\%
& 49.9\% & 43.7\% & 39.3\% \\
Isaac/GR00T N1.5
& 59.2\% & 41.6\% & 29.0\%
& 41.7\% & 24.5\% &  7.6\%
& 24.0\% & 25.4\% & 23.4\% \\
Bimanual/Pi0.5
& 53.3\% & 39.9\% & 22.2\%
& 47.9\% & 28.1\% &  4.1\%
& 24.9\% & 24.3\% & 18.9\% \\

RoboTwin/Fast-WAM
& 57.9\% & 33.0\% & 25.7\%
& 29.0\% & 8.1\% &  8.0\%
& 46.5\% & 32.5\% & 21.7\% \\

Bridge/minic-video
& 52.2\% & 40.1\% & 28.5\%
& 51.0\% & 27.2\% &  2.8\%
& 36.6\% & 31.4\% & 24.5\% \\

\bottomrule
\end{tabular}
\end{subtable}

\vspace{8pt}

\begin{subtable}[t]{\columnwidth}
\subcaption{\small vs.\ Autellix}
\label{tab:latency-reduction-autellix}
\centering
\begin{tabular}{l >{\columncolor{p25color}}c >{\columncolor{avgcolor}}c >{\columncolor{p95color}}c >{\columncolor{p25color}}c >{\columncolor{avgcolor}}c >{\columncolor{p95color}}c >{\columncolor{p25color}}c >{\columncolor{avgcolor}}c >{\columncolor{p95color}}c}
\toprule
& \multicolumn{3}{c}{\name{}} & \multicolumn{3}{c}{\makecell{\name{} w/o \\ dynamic horizon}} & \multicolumn{3}{c}{\makecell{\name{} w/o \\ scheduler}} \\
\cmidrule(lr){2-4}\cmidrule(lr){5-7}\cmidrule(lr){8-10}
Workload
& \tiny P25 & \tiny Avg & \tiny P95
& \tiny P25 & \tiny Avg & \tiny P95
& \tiny P25 & \tiny Avg & \tiny P95 \\
\midrule
LIBERO/SmolVLA
& 41.9\% & 33.1\% & 36.1\%
& 24.2\% & 12.5\% & 11.9\%
&  8.0\% & 15.1\% & 25.5\% \\
LIBERO/XVLA
& 74.9\% & 53.7\% & 46.7\%
& 40.3\% & 21.6\% &  7.1\%
& 43.2\% & 39.4\% & 36.1\% \\
LIBERO/Pi0.5
& 77.4\% & 66.5\% & 52.0\%
& 59.0\% & 34.6\% &  5.9\%
& 55.8\% & 53.9\% & 49.8\% \\
MetaWorld/SmolVLA
& 88.4\% & 63.6\% & 48.2\%
& 63.5\% & 39.3\% & 13.2\%
& 44.5\% & 41.3\% & 36.0\% \\
Isaac/GR00T N1.5
& 61.6\% & 43.3\% & 29.2\%
& 45.2\% & 26.7\% &  7.9\%
& 27.5\% & 24.2\% & 22.9\% \\
Bimanual/Pi0.5
& 55.9\% & 42.3\% & 24.6\%
& 50.8\% & 31.0\% &  7.0\%
& 23.9\% & 23.1\% & 19.8\% \\

RoboTwin/Fast-WAM
& 62.4\% & 36.6\% & 25.2\%
& 36.6\% & 12.9\% &  7.3\%
& 44.2\% & 34.2\% & 22.5\% \\

Bridge/minic-video
& 54.6\% & 41.6\% & 28.7\%
& 53.4\% & 29.1\% &  4.9\%
& 34.9\% & 30.4\% & 25.9\% \\
\bottomrule
\end{tabular}
\end{subtable}
\vspace{-10pt}
\end{table}

\subsection{Trace Fidelity Evaluation with Real Robots}
\label{s:eval_trace_substitution}

When replaying many traces concurrently, scheduling decisions may delay action delivery, causing a robot to stall when freshly generated actions are not yet available; however, once generation completes, the generated actions themselves remain identical to those produced in isolation. In simulation, stalls are benign: the simulator pauses deterministically and resumes from an identical state. We therefore validate through real-robot experiments on the bimanual SO-101 setup that replay-induced stalls do not impact task accuracy. Specifically, we randomly sample 20 tasks from the bimanual online serving experiments (Figure~\ref{fig:aggregate_rate} in \S\ref{s:eval_sched_e2e}) at each arrival rate and inject the corresponding contention-induced delays on the physical robot. Across all arrival rates (0.5, 1.0, and 2.0 tasks/s), accuracy remains stable at 13--14/20, matching the original contention-free trace (13/20). The slight variation at rate\,=\,0.50 is within expected noise from diffusion stochasticity and imperfect initial object placements, confirming the fidelity of our trace-driven evaluation.

\subsection{Sensitivity and Ablation Studies}
\label{s:microbenchmarks}
In this section, we microbenchmark parameters in \name{}'s scheduler. Results are based on three representative workloads; the observed trends hold across all workloads. The experiment setup follows \S\ref{s:eval_setup}.

\begin{figure}[t]
    \centering
    \begin{minipage}[b]{0.48\columnwidth}
        \centering
        \includegraphics[width=\textwidth]{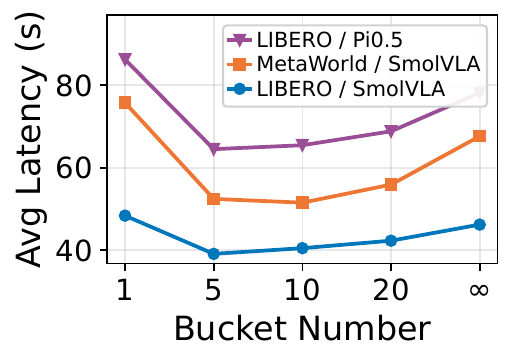}
        \caption{\small \bf Impact of wait ratio bucket count on average latency.}
        \label{fig:microbenchmark-bin-size}
    \end{minipage}
    \hfill
    \begin{minipage}[b]{0.48\columnwidth}
        \centering
        \includegraphics[width=\textwidth]{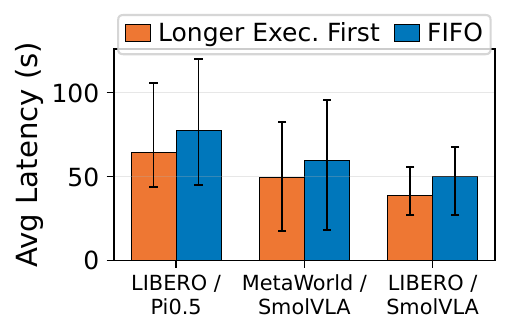}
        \caption{\small \bf Impact of intra-bin sorting policy. Error bars show P25--P95 range.}
        \label{fig:microbenchmark-intra-bin}
    \end{minipage}
    \vspace{-15pt}
\end{figure}

\para{Impact of bucket size.}
Figure~\ref{fig:microbenchmark-bin-size} shows the effect of varying the number of wait ratio buckets (default 10) on average latency. Too few buckets (e.g., 2) coarsen the priority signal, grouping requests with substantially different priorities into the same bucket, and lead to worse latency performance (up to 33.9\% worse). Conversely, too many buckets (e.g., $\infty$, i.e., prioritize by absolute wait ratio) eliminate the stabilizing effect of bucketing, causing frequent switching between scheduling requests. Across all three workloads, 5--10 buckets strike the best balance between priority granularity and scheduling stability.


\para{Impact of intra-bucket policy} Figure~\ref{fig:microbenchmark-intra-bin} validates the choice of sorting by estimated execution latency (descending) within each bucket. Compared to FIFO, it reduces tail latency (P95) by up to 17.9\%, and these per-round gains accumulate into lower average end-to-end latency as well.

\begin{figure}[t]
    \centering
    \includegraphics[width=\columnwidth]{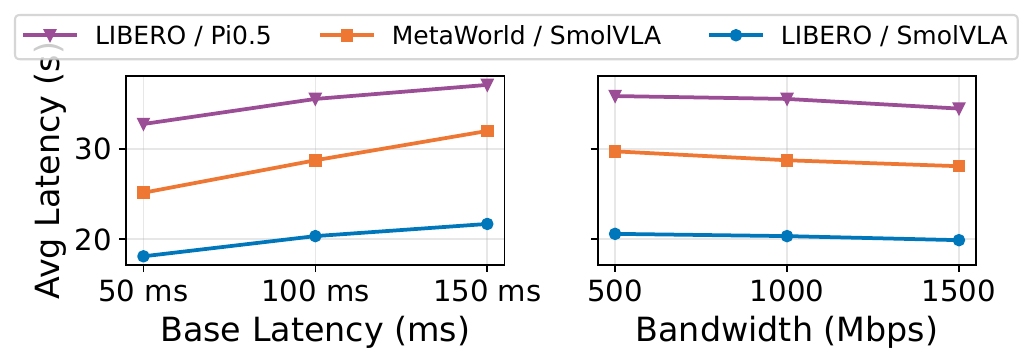}
    \vspace{-10pt}
    \caption{\small \bf Impact of edge to cloud network conditions on average latency. (Left) Varying base latency at 1000\,Mbps bandwidth. (Right) Varying bandwidth at 100\,ms base latency.}
    \label{fig:microbenchmark-network-latency}
    \vspace{-10pt}
\end{figure}

\para{Impact of network latency in hybrid setup.} Figure~\ref{fig:microbenchmark-network-latency} examines how edge-to-cloud network conditions affect \name{}'s hybrid serving performance. Faster network connections (lower base latency, higher bandwidth) benefit \name{} by creating more opportunities to offload requests to the cloud (up to 20.5\% more requests offloaded to cloud). Since observation payloads in our physical AI workloads are small, base latency dominates network latency, making \name{} more sensitive to base latency than bandwidth.

\section{Additional Related Work}
\label{s:related}

\para{LLM and Agent serving.} LLM-based applications have moved beyond text-based chatting toward tool-augmented workflows that interleave token generation
with external function calls~\cite{infercept,gorilla,react}, forming a synchronous loop of generation and execution. Existing works optimize this interleaving through KV-cache management during tool-call pauses to support more concurrent
requests~\cite{infercept,continuum}, but none schedule requests factoring in execution-phase duration, which constitutes a significant fraction of end-to-end latency in physical AI. Other work proposes asynchronous tool
calling~\cite{conveyor,asynclm}, which fully decouples generation from execution so that the model need not wait for tool-call results. Physical AI is fundamentally different: action chunking produces multiple actions per inference call, enabling asynchronous inference without fully decoupling generation from execution; the right time to trigger the next generation depends on how far the current action chunk has been executed. \name{} is built with such execution-awareness for physical AI that digital AI systems lack. 

\para{Efficient physical AI.} Prior work reduces the diffusion cost through skipping diffusion steps~\cite{consistencypolicy,onedp,dream_zero}. Orthogonally, model compression techniques---quantization~\cite{qvla,sqapvla} and pruning~\cite{vlapruner,lightvla}---reduce per-step compute at the model level.
These techniques optimize inference efficiency at the model level and are thus composable with \name{}.

\para{Tiered physical AI systems.}
Several physical AI systems decompose reasoning and control
across tiers: a cloud-hosted LLM performs high-level planning
across multiple tasks while an on-device physical AI model
handles them one by one~\cite{gemini_robotics}.
\name{} focuses on serving the physical AI model tier, treating the upstream planner as an input source. Other systems leverage a dual-system design, pairing a slow perception-reasoning module (System~2) with a fast on-device action generator (System~1)~\cite{figure,gemini_robotics}. \name{}'s techniques seamlessly apply to serving the shared System~2 across robots.

\section{Conclusion}
\label{s:conclusion}

We presented \name{}, the first serving system designed for physical AI serving. \name{} introduces two key mechanisms: a diffusion confidence measure that dynamically adjusts per-round execution horizons, and an execution-aware scheduler that leverages visibility into both generation and execution phases to prioritize requests under contention. Across four model families, three simulation benchmarks, and real-robot experiments, \name{} reduces average end-to-end task latency by 31.8--66.5\% over state-of-the-art serving practices at peak load, scales from 10 to 100 concurrent robots with growing gains, and generalizes to hybrid edge--cloud deployments.
\label{lastpage}
\balance
\Urlmuskip=0mu plus 1mu\relax
\bibliographystyle{abbrv}
\bibliography{paper}

\end{document}